\documentclass[10pt,twocolumn,letterpaper]{article}

\usepackage{cvpr}
\usepackage{times}
\usepackage{epsfig}
\usepackage{graphicx}
\usepackage{amsmath}
\usepackage{amssymb}

\usepackage[sort&compress,numbers]{natbib}
\usepackage{stmaryrd}
\usepackage{makecell}
\usepackage{multirow}
\usepackage{sg-macros}
\usepackage{makecell}
\usepackage{spreadtab}
\usepackage{xcolor}
\usepackage[normalem]{ulem}
\usepackage[ruled,vlined,linesnumbered]{algorithm2e}
\usepackage{enumitem}
\usepackage[misc,geometry]{ifsym} 

\newcommand{\ops}{\mathcal{O}}
\newcommand{\concat}{\succ}
\newcommand{\alt}{\vert}
\newcommand{\kstar}{\star}

\newcommand{\aset}[1]{\left\lbrace #1 \right\rbrace}
\newcommand{\aitem}[1]{a_{#1}}

\usepackage[pagebackref=true,breaklinks=true,letterpaper=true,colorlinks,bookmarks=false]{hyperref}

\cvprfinalcopy 

\def\cvprPaperID{14} 

\ifcvprfinal\fi
\begin{document}

\title{Inferring Temporal Compositions of Actions Using Probabilistic Automata}

 \author{Rodrigo Santa Cruz\textsuperscript{1,2,}\thanks{Corresponding author. Email: rodrigo.santacruz@csiro.au}, Anoop Cherian\textsuperscript{3}, Basura Fernando\textsuperscript{4}, Dylan Campbell\textsuperscript{2}, and Stephen Gould\textsuperscript{2}\\
 \textsuperscript{1}The Australian e-Health Research Centre, CSIRO, Brisbane, Australia\\
 \textsuperscript{2}Australian Centre for Robotic Vision (ACRV), Australian National University, Canberra, Australia\\
 \textsuperscript{3}Mitsubishi Electric Research Labs (MERL), Cambridge, MA\\
 \textsuperscript{4}A*AI, A*STAR Singapore}


\maketitle
\thispagestyle{empty}

\begin{abstract}
This paper presents a framework to recognize temporal compositions of atomic actions in videos. Specifically, we propose to express temporal compositions of actions as semantic regular expressions and derive an inference framework using probabilistic automata to recognize complex actions as satisfying these expressions on the input video features. Our approach is different from existing works that either predict long-range complex activities as unordered sets of atomic actions, or retrieve videos using natural language sentences. 
Instead, the proposed approach allows recognizing complex fine-grained activities using only pretrained action classifiers, without requiring
any additional data, annotations or neural network training.
To evaluate the potential of our approach, we provide experiments on synthetic datasets and challenging real action recognition datasets, such as MultiTHUMOS and Charades.
We conclude that the proposed approach can extend state-of-the-art primitive action classifiers to vastly more complex activities without large performance degradation.
\end{abstract}

\section{Introduction}
\begin{figure*}[t]
	\begin{center}		
		\includegraphics[width=\textwidth]{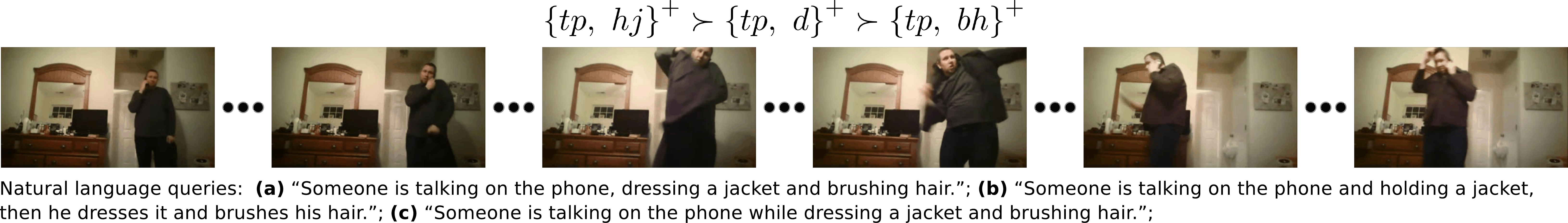}
	\end{center}
	\vspace{-8px}
	\caption{A complex activity can be described by natural language queries, which are often incomplete and have vague and/or ambiguous temporal relations between the constituent actions. For instance, option (a) does not mention all the actions involved, and it is not clear from options (b) and (c) whether the actions happen simultaneously or sequentially. In contrast, a regular expression of primitive actions can precisely describe the activity of interest. For instance, given the primitive actions ``talking on the phone''~(tp), ``holding a jacket''~(hj), ``dressing''~(d), and ``brushing hair''~(bh), the regular expression $\aset{tp, ~hj}^+ \concat \aset{tp,~d}^+ \concat \aset{tp,~bh}^+$ precisely describes the activity depicted in the frames, where the sets of primitive actions $\left(\aset{\cdot}\right)$, the regular language operator `concatenation' ($\concat$) and the operator `one-or-more repetition' ($+$) define concurrent, sequential and recurrent actions, respectively.}
    \label{fig:intro}
    \vspace{-10px}
\end{figure*}

Real-world human activities are often complex combinations of various simple actions. In this paper, we define compositional action recognition as the task of recognizing complex activities expressed as temporally-ordered compositions of atomic actions in videos. To illustrate our task, let us consider the video sequence depicted in \figref{fig:intro}. Our goal is to have this video clip retrieved from a large collection of videos. A natural query in this regard can be: \emph{``find videos in which someone is holding a jacket, dressing, and brushing hair, while talking on the phone?''}. As is clear, this query combines multiple atomic actions such as ``holding a jacket'', ``dressing'', etc.; however, we are interested only in videos that adhere to the temporal order provided in the query. Such a task is indispensable in a variety of real-world applications, including video surveillance~\citep{Nouby:PhDThesis16}, patient monitoring~\citep{Lau:2010}, and shopping behavior analysis~\citep{Liciotti:2014}.


The current state-of-the-art action recognition models are usually not equipped to tackle this compositional recognition task. Specifically, these models are often trained to either discriminate between atomic actions in a multi-class recognition setting -- such as ``holding a jacket'', ``talking on the phone'', etc.~\citep{Tran:ICCV15,Carreira:CVPR2017}, or  generate a set of action labels -- as in a multi-label classification setup -- for a given clip, e.g., $\{$``cooking a meal'', ``preparing a coffee''$\}$~\citep{Wang:CVPR18,Hussein:CVPR19,Hussein:Arxiv20}. On the one hand, extending the multi-class framework to complex recognition may need labels for all possible compositions -- the annotation efforts for which could be enormous. On the other hand, the multi-label formulation is designed to be invariant to the order of atomic actions and thus is sub-optimal for characterizing complex activities.


Recent methods attempt to circumvent these limitations by leveraging textual data, allowing zero-shot action recognition from structured queries expressed as natural language sentences.
For instance, these are models able to perform zero-shot action classification \citep{Jain:ICCV2015,Mettes:ICCV2017}, action localization \citep{Gao:ICCV2017,Hendricks:ICCV2017,Liu:ECCV18}, and actor-action segmentation \citep{Gavrilyuk:CVPR2018} from natural language sentences.
However, such descriptions are inherently ambiguous and not suitable for precisely describing complex activities as temporal compositions of atomic actions. As an example, all natural language descriptions listed for the video shown in \figref{fig:intro} are true, but none of them describe the sequence of events precisely. 

Building on the insight that complex activities are fundamentally compositional action patterns, we introduce a probabilistic inference framework to \emph{unambiguously} describe and efficiently recognize such activities in videos. To this end, we propose to describe complex activities as regular expressions of primitive actions using \emph{regular language operators}. Then, using probabilistic automata~\citep{Rabin:1963}, we develop an inference model that can recognize these regular expressions in  videos. Returning to the example in \figref{fig:intro}, given the primitive actions ``talking on the phone''~(tp), ``holding a jacket''~(hj), ``dressing''~(d), and ``brushing hair''~(bh), the regular expression $\aset{tp, ~hj}^+ \concat \aset{tp,~d}^+ \concat \aset{tp,~bh}^+$ precisely describes the activity in the video, where the sets of primitives $\left(\aset{\cdot}\right)$, the regular language operators `concatenation' ($\concat$), and `one-or-more repetition' operator ($+$) define concurrent, sequential, and recurrent actions respectively. 

Our framework offers some unique benefits. It can recognize rare events, such as an ``Olympic goal'' for example, by composing ``corner kick'', ``ball traveling'', and ``goal''. Further, it does not require any additional data or parameter learning beyond what is used for the primitive actions. Thus, it can scale an existing pretrained action recognition setup towards complex and dynamic scenarios.

\noindent Below we summarize our main contributions:
\begin{enumerate}[nolistsep]
\item We propose to recognize complex activities as temporal compositions of primitive action patterns. We formulate a framework for this task that resembles a regular expression engine in which we can perform inference for any compositional activity that can be described as a regular expression of primitives.
\item Using probabilistic automata, we derive a model to solve the inference problem leveraging pretrained primitive action classifiers, without requiring additional data, annotations or parameter learning. 
\item Apart from experiments on a synthetic dataset, we evaluate the proposed framework on the previously described task of compositional action recognition using trimmed and untrimmed videos from challenging benchmarks such as MultiTHUMOS~\citep{Yeung:IJCV2015} and Charades~\citep{Sigurdsson:ECCV2016}. 
\end{enumerate}

\section{Related Work}
The action recognition community has mostly focused on developing models to discriminate between short and simple actions~\citep{Kang:2016,Herath:2017}. While getting very accurate in this context, these models require training data for every action of interest which is often infeasible.
Zero-shot learning approaches~\citep{Lampert:PAMI2014,Liu:CVPR2011,Jain:ICCV2015,Habibian:PAMI2017} were developed to mitigate this problem, but these models still interpret action recognition as the assignment of simple action labels. 
In contrast, we follow a compositional view of action recognition where complex actions are inferred from simple primitive actions.

Recently, the community has shifted focus to the recognition of the long-tailed distribution of complex activities.
However, existing methods~\citep{Hussein:CVPR19,Hussein:Arxiv20,Wang:CVPR18} tackle this problem in a multi-label classification setup where the goal is to predict the atomic actions necessary to accomplish a complex activity under weak and implicitly-defined temporal relationships. In contrast, our proposed framework allows querying complex activities with specific and explicitly provided temporal order on the atomic actions.

Another approach to recognize complex activities with partial temporal structure is to leverage vision-language models~\citep{Gao:ICCV2017,Hendricks:ICCV2017,Gavrilyuk:CVPR2018,Liu:ECCV18}.
We argue, however, that natural language sentences may lead to ambiguous descriptions of complex activities as shown in \figref{fig:intro}, which makes it difficult to ground the videos and their queries. Instead, we resort to regular languages that are designed to unambiguously describe concepts and also allows efficient inference.

Serving as inspirations for our approach, the works of \citet{Ikizler:IJCV2008} and \citet{Vo:CVPR2014} recognize human-centered activities from compositions of primitive actions. 
However, these approaches can only query for sequential or alternative primitive actions of \emph{fixed length}. 
In contrast, we propose a more expressive and complete language for querying complex activities allowing sequential, concurrent, alternative, and recursive actions of varying lengths.
Furthermore, our work focuses on zero-shot recognition of complex activities, unlike these approaches which require training data for the queried activities.

Note that our work is different from ones that leverage manually-annotated training data to perform structured predictions involving primitive actions. 
For instance, \citet{Richard:CVPR2016} and \citet{Pier:CVPR18} consistently label video frames in a sequence, while Ji et al.'s model~\citep{Ji:Arxiv19} outputs space-time scene graphs capturing action-object interactions. 
Differently, we tackle zero-shot activity classification over a highly complex label space (\ie, the space of all regular expressions of primitive actions) by using a probabilistic inference framework that uses only pretrained primitive action classifiers and does not need training data for action composites, or classifier finetuning.
\section{Approach}
In this section, we first formalize the problem of recognizing complex activities described by regular expressions of primitive actions. 
Then, we derive a deterministic and probabilistic inference procedure for this problem.

\subsection{Problem Formulation}
Initially, let us assume the existence of a pre-defined set of known actions, called primitives.
For example, ``driving''~(d), ``getting into a car''~(gc), ``talking on the cellphone''~(tc), and ``talking to someone''~(ts).
These primitives can also happen simultaneously in time, which we express as subsets of these primitive actions, \eg, $\aset{\aitem{d},~\aitem{tc}}$ means someone driving and talking on the cellphone at the same time.
Moreover, consider three basic composition rules inspired by standard regular expression operators: concatenation $\left( \concat \right)$, alternation $(\alt)$, and Kleene star $(\kstar)$ denoting sequences, unions and recurrent action patterns, respectively. 
Note also that more complex operators can be defined in terms of these ones, \eg, one-or-more repetition of action ($+$) is defined as $ a^+ \triangleq a \concat a^\kstar$.
Then, from any complex activity described as a composition of subsets of primitive actions and these operators, our goal is to recognize whether a given video depicts the described activity.
For instance, whether a given YouTube video depicts the complex activity ``someone driving and talking on the phone or talking to someone, repeatedly, just after getting into a car'', which can be described unambiguously as ``$\aitem{gc} \concat \left(\aset{\aitem{d}, ~\aitem{tc}} \alt \aset{\aitem{d},~\aitem{ts}}\right)^+$''.

Formally, let us define a set of \emph{primitive actions} $\A=\{\aitem{i}\}_{i=1}^M$. We can express a complex activity by forming \emph{action patterns}, an arbitrary regular expression $r$ combining subsets of primitives $w \in \P\left(\A\right)$, where $\P\left(\A\right)$ is the power-set of $\A$, with the aforementioned \emph{composition rules} $\ops =\{ \concat, \alt, \kstar \}$. Note that background actions and non-action video segments are represented by the null primitive $\emptyset \in \P\left(\A\right)$. Our goal then is to model a function $f_r: \V \rightarrow \left[ 0, 1 \right]$ that assigns high values to a video $v \in \V$ if it depicts the action pattern described by the regular expression $r$ and low values otherwise. 

This work focuses on solving the aforementioned problem by developing a robust inference procedure leveraging state-of-the-art pretrained primitive action classifiers. Such an approach does not require extra data collection, parameter learning or classifier finetuning. 
Indeed, learning approaches are beyond the scope of this paper and a compelling direction for future work.

\subsection{Deterministic Model}
Regular expressions are used to concisely specify patterns of characters for matching and searching in large texts~\citep{Lawson:2003,Mitkov:2003,Sedgewick:2011}. Inspired by this idea, we now describe a deterministic model based on deterministic finite automata (DFA)~\citep{McCulloch:1943,Rabin:1959} to the problem of recognizing action patterns in videos. 

Let us start by defining a DFA $M_r$ for an action pattern $r$ as a 5-tuple $\left(\Q, \Sigma, \delta, q_{0}, \F \right)$, consisting of a finite set of states $Q$, a finite set of input symbols called the alphabet $\Sigma$, a transition function $\delta :\Q \times \Sigma \rightarrow \Q$, an initial state $q_0 \in \Q$, and a set of accept states $\F \subseteq \Q$. In our problem, the alphabet $\Sigma$ is the power-set of action primitives $\P(\A)$ and the transition function $\delta$ is a lookup table mapping from a state $q_i \in \Q$ and a subset of primitives $w \in \Sigma$ to another state $q_j \in \Q$ or halting the automaton operation if no transition is defined.
Note that all these structures are efficiently constructed and optimized from a given action pattern $r$ using traditional algorithms such as non-deterministic finite automaton (NFA) construction~\citep{Ilie:2002}, the NFA to DFA subset construction~\citep{Rabin:1959}, and Hopcroft's DFA minimization~\citep{Hopcroft:1971}.

Additionally, let us denote the subset of primitive actions happening in a given frame $x$ as ${w(x)=\left\lbrace a \in \A \mid p(a \vert x) \geq \tau \right\rbrace}$, where $p(a \vert x)$ is the probability of a primitive action $a \in \A$ happening in frame $x$ and $\tau \in \left[0, 1\right]$ is a threshold hyper-parameter. In this formulation, $p(a \vert x)$ can be built from any probabilistic action classifier, while  $\tau$ should be set by cross-validation. Then, we say that the deterministic model accepts an input video $v=\langle x_1, \ldots, x_n \rangle$ if and only if there exists a sequence of states $\langle q_0, \ldots, q_n \rangle$ for  $q_i \in \Q$ such that (i) the sequence starts in the initial state $q_0$, (ii) subsequent states $q_i$ satisfy $q_{i} = \delta\left(q_{i-1}, w\left(x_{i}\right)\right)$ for $i=1, \ldots, n$, and (iii) the sequence finishes in a final state $q_n \in \F$.

This procedure defines a binary function that assigns a value of one to videos that reach the final state of the compiled DFA $M_r$ and zero otherwise. This is a very strict classification rule since a positive match using imperfect classifiers is very improbable. In order to relax such a classification rule, we propose implementing the score function
\begin{equation}
f_r(v) = \frac{\dist(q_0, \hat{q})}{\dist(q_0, \hat{q}) + \min_{q_f \in \F} \dist\left(\hat{q},  q_f \right)} ~,
\label{eq:det_soft_score}
\end{equation}
where $\hat{q}$ is the state in which the compiled DFA $M_r$ halted when simulating the sequence of frames defined by the video $v$, and the function $\dist(q_x, q_y)$ computes the minimum number of transitions to be taken to reach the state $q_y$ from state $q_x$. That is, for a given regular expression, the deterministic model scores a video according to the fraction of transitions taken before halting in the shortest path to a final state in the compiled DFA. 

In short, the deterministic model implements the function $f_r$ by computing \eqnref{eq:det_soft_score} after simulating the DFA $M_r$ compiled for the regular expression $r$ on the sequence of subsets of primitive actions $w(x)$ generated by thresholding the primitive action classifiers $p(a \vert x)$ on every frame $x$ of the input video $v$.

\subsection{Probabilistic Model}
In order to take into account the uncertainty of the primitive action classifiers' predictions, we now derive a probabilistic inference model for our problem. Specifically, we propose to use Probabilistic Automata (PA) \citep{Rabin:1963} instead of DFAs as the backbone of our framework. 

Mathematically, let us define a PA $U_r$ for a regular expression $r$ as a 5-tuple $(\Q, \Sigma, T, \brho, \F)$.
$\Q$, $\Sigma$, and $\F$ are the set of states, the alphabet, and the final states, respectively. 
They are defined as in the deterministic case, but an explicit reject state is added to $\Q$ in order to model the halting of an automaton when an unexpected symbol appears in any given state.
${T = \lbrace \forall w \in \Sigma: T_w \in \reals^{\len{\Q} \times \len{\Q}} \rbrace}$ is the set of row stochastic transition matrices $T_w$ associated with the subset of primitives $w \in \Sigma$ in which the entry $\left[ T_{w} \right]_{i,j}$ is the probability that the automaton transits from the state $q_i$ to the state $q_j$ after seeing the subset of primitives $w$. Likewise, $\brho \in \reals^{\len{\Q}}$ is a stochastic vector and $\left[\brho\right]_i$ is the probability that the automaton starts at state $q_i$.

Note that all these structures are estimated from the transition function $\delta$ and initial state $q_0$ of the compiled DFA $M_r$ for the same regular expression $r$ as follows,
\begin{equation}
\begin{aligned}
\left[T_w\right]_{i,j} &= \frac{\ind{\delta(i, w) = j} + \alpha}{\displaystyle\sum_{k \in \Q}\ind{\delta(i, w) = k}  + \alpha\len{\Q}}\,, \\
\left[\brho\right]_i &= \frac{\ind{q_0 = i} + \alpha}{\displaystyle\sum_{k \in \Q}\ind{q_0 = k}  + \alpha\len{\Q}}\,,
\label{eq:tran_dist}
\end{aligned}
\end{equation}
where the indicator function $\ind{c}$ evaluates to one when the condition $c$ is true and zero otherwise. The smoothing factor $\alpha$ is a model hyper-parameter that regularizes our model by providing non-zero probability for every distribution in our model. As mentioned before, a hypothetical dataset of action patterns and videos pairs could be leveraged by a learning algorithm to fit these distributions, but the current work focuses on the practical scenario where such a training set is difficult to obtain.


However, PAs do not model uncertainty in the input sequence which is a requirement of our problem, since we do not know what actions are depicted in a given video frame. Therefore, we propose to extend the PA framework by introducing a distribution over the alphabet given a video frame. In order to make use of off-the-shelf action classifiers like modern deep leaning models, we assume independence between the primitive actions and estimate the probability of a subset of primitives given a frame as
\begin{equation}
    p(w \vert x) = \left(  \displaystyle\prod_{a \in \A} p(a \vert x)^{\ind{a \in w}} \left(1 -  p(a \vert x)\right)^{\ind{a \notin w}} \right)^{\!\gamma},
\end{equation}
where $p(a \vert x)$ is the prediction of a primitive action classifier as before and $\gamma$ is a hyper-parameter that compensates for violations to the independence assumption. After such a correction, we need to re-normalize the $p(w \vert x)$ probabilities in order to form a distribution. 

Making use of this distribution, we derive the induced (row stochastic) transition matrix $I(x) \in \reals^{\len{\Q} \times \len{\Q}}$ after observing a video frame $x$ by marginalizing over the alphabet $\Sigma$ as follows,
\begin{equation}
I(x) = \sum_{w \in \Sigma} T_w p(w \mid x),
\label{eq:ind_tran_dist}
\end{equation}
where the entry $\left[I(x)\right]_{i,j}$ denotes the probability of transiting from state $q_i$ to state $q_j$ after seeing a frame $x$. It is also important to note that naively computing this induced transition matrix is problematic due to the possibly large alphabet $\Sigma$. For instance, a modestly sized set of $100$ primitive actions would generate an alphabet of $2^{100}$ subsets of primitives. In order to circumvent such a limitation, we factorize \eqnref{eq:ind_tran_dist} as
\begin{equation}
    I(x) = \sum_{w \in \Sigma^{\prime}} T_w p(w \vert x) + \bar{T} \Big(1 - \sum_{w \in \Sigma^{\prime}} p(w \vert x)\Big),
\label{eq:fac_prob_match}
\end{equation}
where we first define a typically small subset of our alphabet $\Sigma^{\prime} \subseteq \Sigma$ composed of subsets of primitives that have at least one transition in the compiled DFA $M_r$. Then, we make use of the fact that the remaining subsets of primitives $\Sigma \setminus \Sigma^{\prime}$ will be associated with exactly the same transition matrix, denoted by $\bar{T}$ and also computed according to \eqnref{eq:tran_dist}, and the sum of their probability in a given frame is equal to $1 - \sum_{w \in \Sigma^{\prime}} p(w \mid x_i)$. 
Therefore, \eqnref{eq:fac_prob_match} computes the induced transition matrix efficiently, without enumerating all subsets of primitives in the alphabet.

Finally, we can compute the normalized matching probability between a video $v=\langle x_1, \ldots x_n \rangle$ and the regular expression $r$ as the probability of reaching a final state in the compiled PA $U_r$ as
\begin{equation}
P_{U_r}\left(v\right) = \left(\brho\transpose\prod_{i=1}^{|v|}I(x_i)\right)^{\!\frac{1}{\len{v}}} \bbf\,,
\label{eq:prob_match}
\end{equation}
where $\bbf$ is an indicator vector such that $\bbf_i=1$ if and only if $q_i \in \F$ and $0$ otherwise. The normalization by $1/\len{v}$ calibrates the probabilities to allow comparisons between videos of different length. 

Intuitively, the proposed probabilistic inference model implements the function $f_r$ by first converting the compiled DFA $M_r$, for the regular expression $r$, to a PA $U_r$ according to \eqnref{eq:tran_dist}. Then, as described in \eqnref{eq:prob_match}, this model keeps a distribution over the states $\Q$ starting from the initial state distribution $\brho$ and updating it according to the induced transition matrix $I(x)$, defined in \eqnref{eq:fac_prob_match}, as we observe the input video frames $x$. Finally, as also described in \eqnref{eq:prob_match}, the matching probability is computed as the sum of the probability in the final states once all of the input video frames are observed.

\section{Experiments}
We now evaluate the proposed inference models for rich compositional activity recognition. We first perform a detailed analysis of the proposed approaches on controlled experiments using synthetic data. Then, we test the utility of our methods on challenging action recognition tasks.

\subsection{Synthetic Analysis \label{sec:mnist}}
\begin{figure}[t!]
	\begin{center}
	\includegraphics[width=0.27\textwidth]{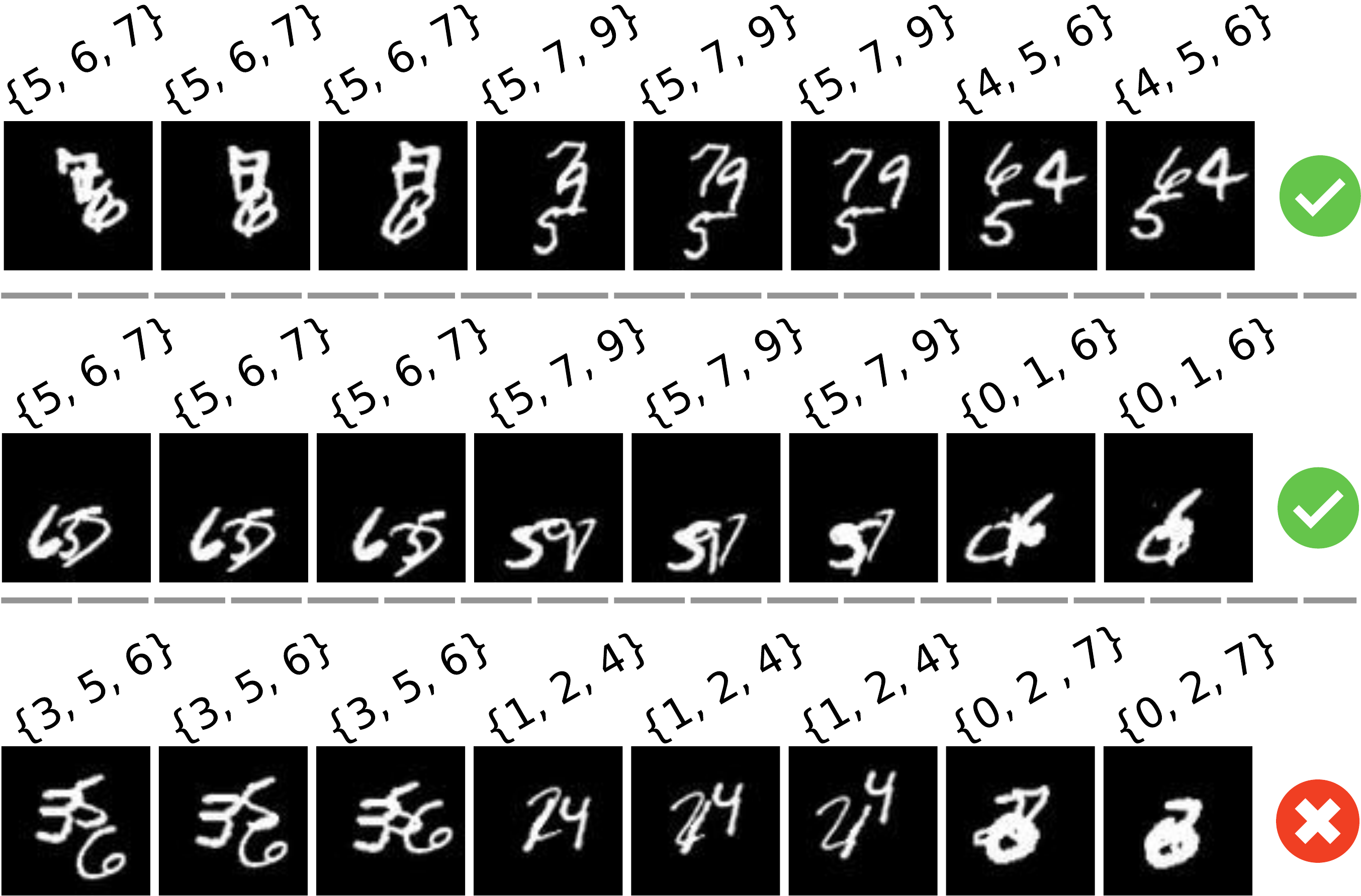}
	\hfill\vline\hfill
	\includegraphics[width=0.20\textwidth, trim={0 0 11.5cm 0},clip]{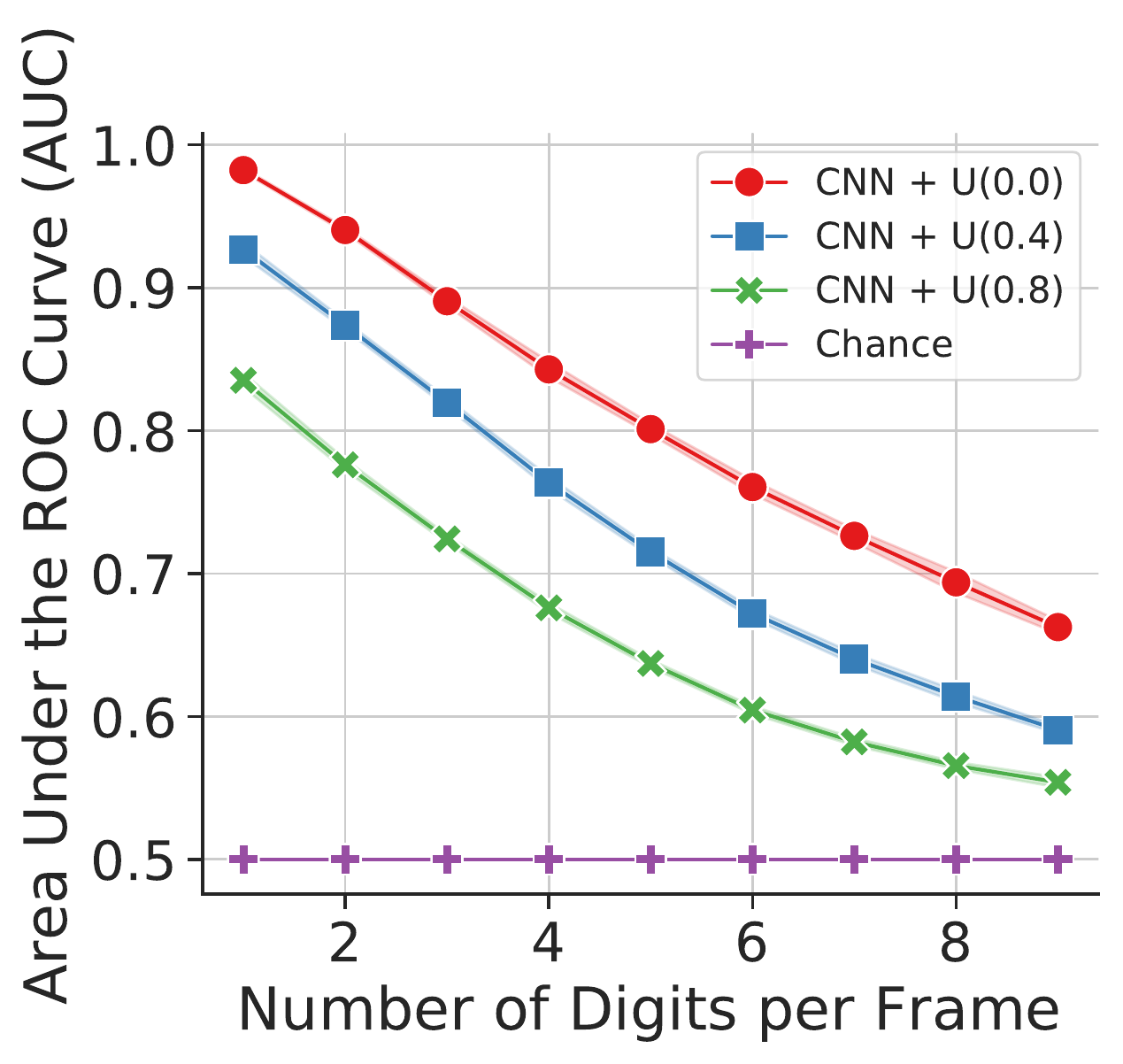}
	\end{center}
\vspace{-8px}
\caption{\textbf{Left:} Video samples synthetically generated for the regular expression  ``$\aset{\aitem{5},~\aitem{6},~\aitem{7}}^+ \concat \aset{\aitem{5},~\aitem{7},~\aitem{9}}^+ \concat  \left(\aset{\aitem{4},~\aitem{5},~\aitem{6}}^+ \alt \aset{\aitem{0},~\aitem{1},~\aitem{6}}^+\right)$'' which has $\vert w \vert=3$ digits per frame, $n=3$ different sequential symbols, $d=2$ alternative sequences starting from $s=2$, depicted in a total of $8$ frames. The two first rows are clips that match the given regular expression, while the last row depicts a negative video clip. \textbf{Right:} The performance of the primitive classifiers on the test set with a different number of digits per frame and under different noise levels. $U(x)$ denotes uniform additive noise between $[-x, x]$ and the classifiers' predictions are re-normalized using a softmax function.}
\label{fig:mnist_data_cls}
\vspace{-10px}
\end{figure}

It is unrealistic to collect video data for the immense number of possible regular expressions that our models may encounter. As such, we resort to the use of synthetically generated data inspired by the well known Moving MNIST dataset \citep{Srivastava:ICML2015}. More specifically, we develop a parametrized data generation procedure to produce moving MNIST videos depicting different patterns of appearing MNIST digits. Such a procedure can generate videos that match regular expressions of the form
\begin{equation}
\resizebox{.89\hsize}{!}{$w_1^+ \succ \cdots \\ \succ \left(\left({w_s^1}^+ \succ \cdots \succ {w_n^1}^+ \right) \bigg\vert \cdots \bigg\vert \left({w_s^d}^+ \succ \cdots \succ {w_n^d}^+\right) \right)$},
\label{eq:mnist_regex}
\end{equation}
where the symbols $w \in \P(\A)$ are subsets of the primitives $\A$ which are the ten digit classes. The data generation procedure has the following parameters: the number of primitives that simultaneously appear in a frame $\vert w\vert$, the total number of different sequential symbols $n$, the number of alternative sequences of symbols $d$, the start position $s$ of each alternative sequence in the pattern, and the total number of generated frames. Since complex patterns can match different sequences of symbols due to the the alternation operator $(\vert)$, we perform random walks from the start state until reaching a final state in the compiled DFA in order to generate video samples for a given regular expression. \figref{fig:mnist_data_cls} presents an example of regular expressions and video clips generated by this data generation procedure.

\begin{figure*}[t!h]
	\begin{center}
		\includegraphics[width=\textwidth]{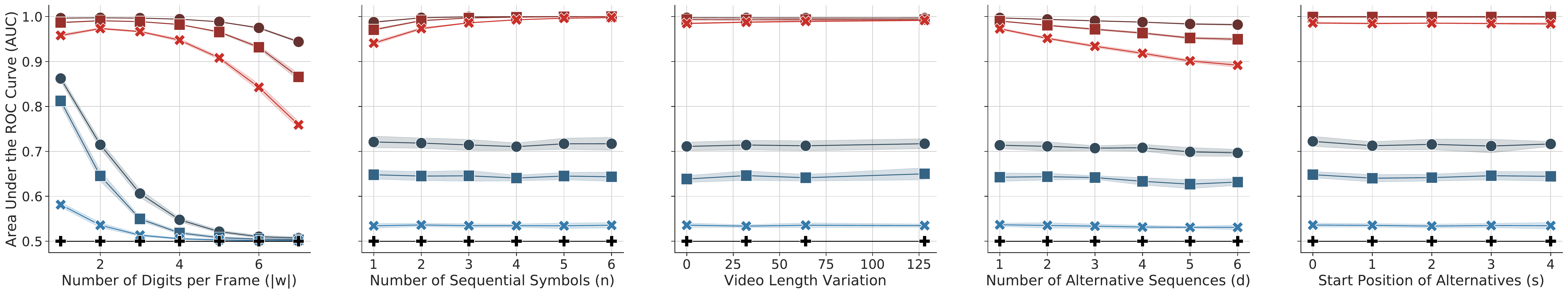}
		\includegraphics[width=\textwidth]{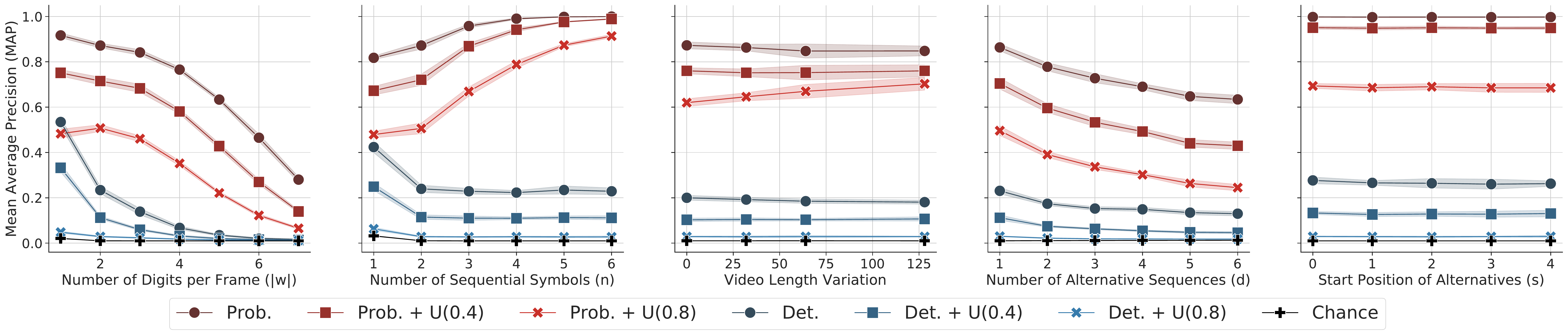}
	\end{center}
\vspace{-8px}
\caption{Plots of the performance, in terms of AUC and MAP, of the proposed methods on the generated synthetic dataset using primitive classifiers with different levels of noise as shown in \figref{fig:mnist_data_cls}. The generated data consists of video clips depicting regular expressions parametrized according to \eqnref{eq:mnist_regex}. We evaluate the proposed approaches according to the following data parameters: the number of digits that simultaneously appear in a frame $\vert w\vert$, the total number of different sequential symbols $n$, the variance in number of frames in the videos, the number of alternative sequences of symbols $d$, and the start position $s$ of each alternative sequence in the pattern respectively.}
\label{fig:mnist_exp}
\vspace{-10px}
\end{figure*}

Using the synthetically generated data, we first train the primitive classifiers on frames depicting a different number of digits obtained from the MNIST training split. The primitive classifiers consist of a shallow CNN trained to minimize the binary cross entropy loss for all digits in a vast number of frames. In order to evaluate the robustness of the proposed models, we also generate worse versions of these classifiers by adding noise to their predictions. \figref{fig:mnist_data_cls} shows the performance of the learned primitive classifiers on different levels of noise and different numbers of digits per frame. Note that more digits per frame implies more occlusion between digits since the frame size is kept constant, which also decreases the classifier's performance.

Finally, using this synthetic data and the trained primitive classifiers, we test our models for the inference of different regular expressions by setting all the data generation parameters to default values with the exception of the one being evaluated. We use the values described in \figref{fig:mnist_data_cls} as default values, but we generate video clips of 32 frames. In \figref{fig:mnist_exp}, we plot standard classification/retrieval metrics, e.g., Area Under the ROC Curve (AUC) and Mean Average Precision (MAP), against different data generation parameters. More specifically, at each configuration, using the MNIST test split, we generate 100 expressions with 20 positive samples, totaling about 2000 video clips. In order to robustly report our results, we repeat the experiment ten times reporting the mean and standard deviation of the evaluation metrics. We also cross-validate the model hyper-parameters, $\tau$ for the deterministic model and $\alpha$ and $\gamma$ for the probabilistic model, in a validation set formed by expressions of similar type as the ones to be tested, but with video clips generated from a held-out set of digit images extracted from the training split of the MNIST dataset.

As can be seen, the probabilistic model performs consistently better than the deterministic model in all experiments, providing precise and robust predictions. Furthermore, the probabilistic model is more robust to high levels of noise in the primitive classifiers' predictions. While the deterministic model works as poorly as random guessing with high noise levels, \eg $U(0.8)$, the probabilistic model still produces good results.
The probabilistic model also works consistently across different kinds of regular expressions. 
Indeed, its performance is almost invariant to most of the regular expressions parameters evaluated.
In the case of the number of digits per frame $\vert w \vert$ for which relevant performance degradation is observed, the performance degradation correlates with the decrease in performance presented by the primitive classifiers as the number of digits per frames is increased (see \figref{fig:mnist_data_cls}). 
The probabilistic model, however, is able to mitigate this degradation.
For example, comparing the performance at two and five digits per frame, we observe that a drop of about 16\% in AUC on the primitive classifiers performance causes a reduction smaller than 3\% in AUC for the probabilistic model. 

\begin{figure*}[t!h]
	\begin{center}
	    \begin{tabular}{c|c}
            \footnotesize{Expressions mined from MultiTHUMOS} & \footnotesize{Expressions mined from Charades} \\
           	\includegraphics[width=0.49\textwidth,trim={0 0 0 1.15cm},clip]{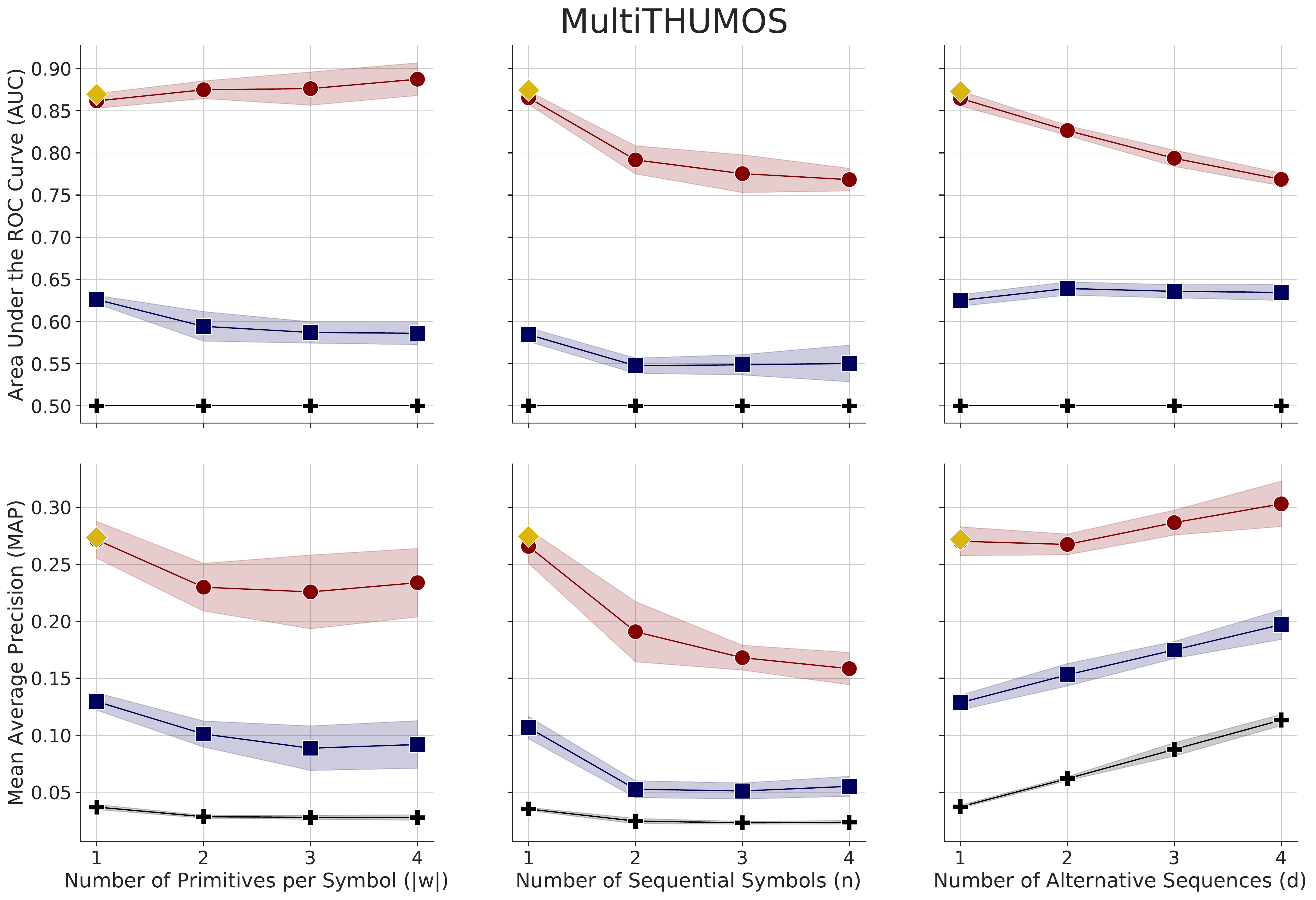} & \includegraphics[width=0.49\textwidth,trim={0 0 0 1.15cm},clip]{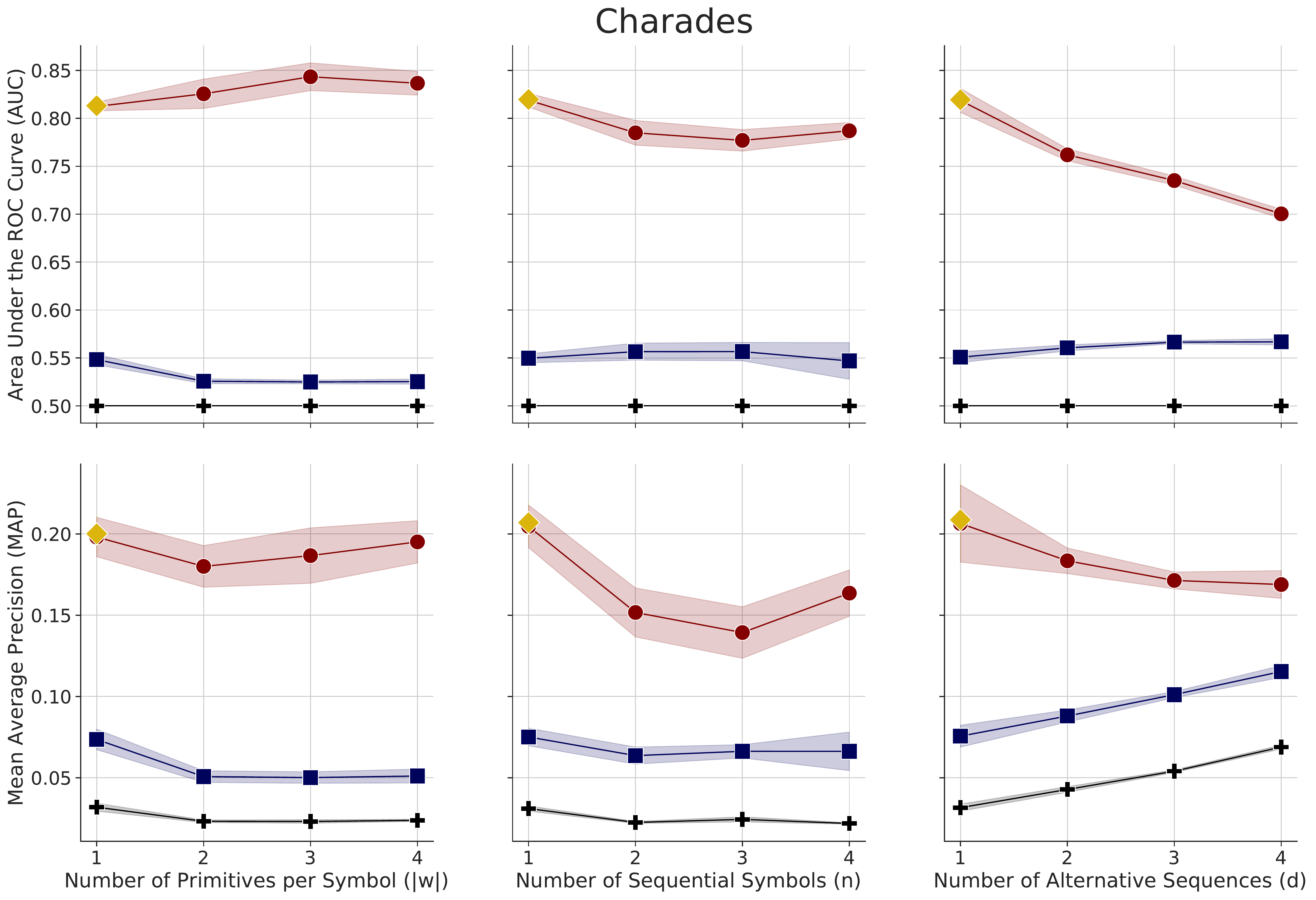} \\
        \end{tabular}
		\includegraphics[width=0.8\textwidth,trim={0 0.5cm 0 31.5cm},clip]{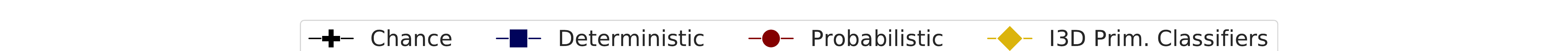}
	\end{center}
\vspace{-8px}
\caption{Comparison with standard action classification. Plots of the performance, in terms of AUC and MAP, of the proposed methods using the I3D model \citep{Carreira:CVPR2017} as the primitive action classifier. We evaluate the models on collections of regular expressions of different complexity mined from the test videos of MultiTHUMOS and Charades datasets. These regular expressions follows the format defined in \eqnref{eq:mnist_regex} where all the variables are set to 1 with the exception of the one being evaluated. For instance, for the plot with variable number of sequential symbols $(n)$ the expressions vary from $(w_1^+)$ to $(w_1^+ \concat \cdots \concat w_4^+)$. Differently from the other experiments, the symbols here denote any subset that contains the primitives.}
\label{fig:res_ar_com}
\vspace{-10px}
\end{figure*}

\subsection{Action Recognition}
We now evaluate our models on action recognition problems. We first describe the experimental setup, metrics and datasets used in our experiments. Then we analyze how effectively our model can recognize activities described by regular expressions in trimmed and untrimmed videos.  

\noindent\textbf{Experimental Setup.} In order to evaluate the proposed inference models in the action recognition context, we collect test datasets of regular expressions and video clips by mining the ground-truth annotation of multilabel action recognition datasets such as Charades~\citep{Sigurdsson:ECCV2016} and MultiTHUMOS~\citep{Yeung:IJCV2015}. More specifically, we search for regular expressions of the type defined in \eqnref{eq:mnist_regex} where the symbols $w$ are subsets of the primitive actions annotated in the datasets. Charades has 157 actions, while MultiTHUMOS has 65 actions. Given the regular expression parameters, we first form instances of regular expressions using the primitive actions present in the datasets, keeping the ones that have at least one positive video clip. Then, using these instances of regular expressions, we search for all positive video clips in the dataset in order to form a new dataset of regular expressions and video clips which will be used in our experiments. 

As primitive action classifiers, we use the I3D model proposed by \citet{Carreira:CVPR2017} pretrained on the Kinetics dataset and finetuned on the training split of the Charades and MultiTHUMOS datasets to independently recognize the primitive actions. In this work, we only use the I3D-RGB stream, but optical flow and other information can be easily added since our formulation depends only on the final predictions of the primitive classifiers. Using the frame-level evaluation protocol (\ie Charades localization setting), these classifiers reach \textbf{16.12\% and 24.93\% in MAP} on classifying frames into primitive actions on the test split of Charades and MultiTHUMOS datasets respectively.

Once the primitives classifiers are defined, we setup the deterministic and probabilistic inference models with them, cross-validate these inference model hyper-parameters using expressions and video clips mined from the training split, and evaluate these inference models in a different set of expressions mined from the test split of the aforementioned action recognition datasets. 
It is important to emphasize that the expressions mined for testing are \emph{completely} different from the ones used for cross-validation.
Therefore, the proposed inference models have not seen any test frame or the same action pattern before, which provides an unbiased evaluation protocol. In order to provide robust estimators of performance, in the experiments of the current section, we repeat the data collection of 50 regular expressions and the test procedure steps ten times, reporting the mean and standard deviation of the evaluation metrics AUC and MAP. Note that these metrics are computed over the recognition of the \emph{whole complex activity} as a singleton label. They are \emph{not} computed per primitive.

\noindent\textbf{Comparison To Standard Action Recognition.} Traditional action classification aims to recognize a single action in a video, making no distinction if the action is performed alone or in conjunction with other actions. Abusing the proposed regular expression notation, for now consider the symbols $w$ in \eqnref{eq:mnist_regex} as the collection of all subsets of the primitive actions that contains the actions in $w$. For instance, the symbol $\aset{\aitem{2},\aitem{3}}$ represents (only here) the set of symbols $\aset{\aset{\aitem{2},\aitem{3}}, \aset{\aitem{2},\aitem{3}, \aitem{4}}, \ldots  \aset{\aitem{2},\aitem{3}, \aitem{4}, \ldots, \aitem{\len{\A}}}}$. Then, we can say that the traditional action classification problem is the simplest instance of our formulation where the input regular expressions are of the type $\aset{\aitem{}}^+$, meaning one or more frames depicting the action $a$ alone or in conjunction with other actions. Therefore, starting from this simplified setup, we analyze how our models behave as we increase the difficulty of the problem by dealing with more complex regular expressions. More specifically, we start from this simplest form, where all the regular expression parameters are set to one, and evolve to more complex expressions by varying some of the parameters separately. \figref{fig:res_ar_com} presents the results on expressions mined from the Charades and MultiTHUMOS datasets where we vary the number of concurrent (columns 1 and 4), sequential (columns 2 and 5), and alternated actions (columns 3 and 6) by varying the number of primitives per symbol $\len{w}$, the number of sequential symbols $n$, and the number of alternative sequences $d$ in the mined regular expression and video clip data, respectively.

Note that there is a significant difference in performance when compared to the results in \secref{sec:mnist}. Such a difference is due to the quality of the primitive classifiers available for a challenging problem like action classification. For instance, the digits classifiers for the MNIST dataset are at least three times more accurate than the primitive action classifiers for Charades or MultiTHUMOS. However, different from the deterministic model, the probabilistic model is able to extend the primitive action classifiers, the I3D model, for complex expressions without degenerating the performance significantly. For instance, considering all setups, the probabilistic model presents a reduction in performance of at most 15\% in both datasets and metrics used. This result suggests that the proposed model can easily scale up the developments in action recognition to this challenging compositional activity recognition problem.


\begin{figure*}[t!]
	\begin{center}		
		\includegraphics[width=\textwidth]{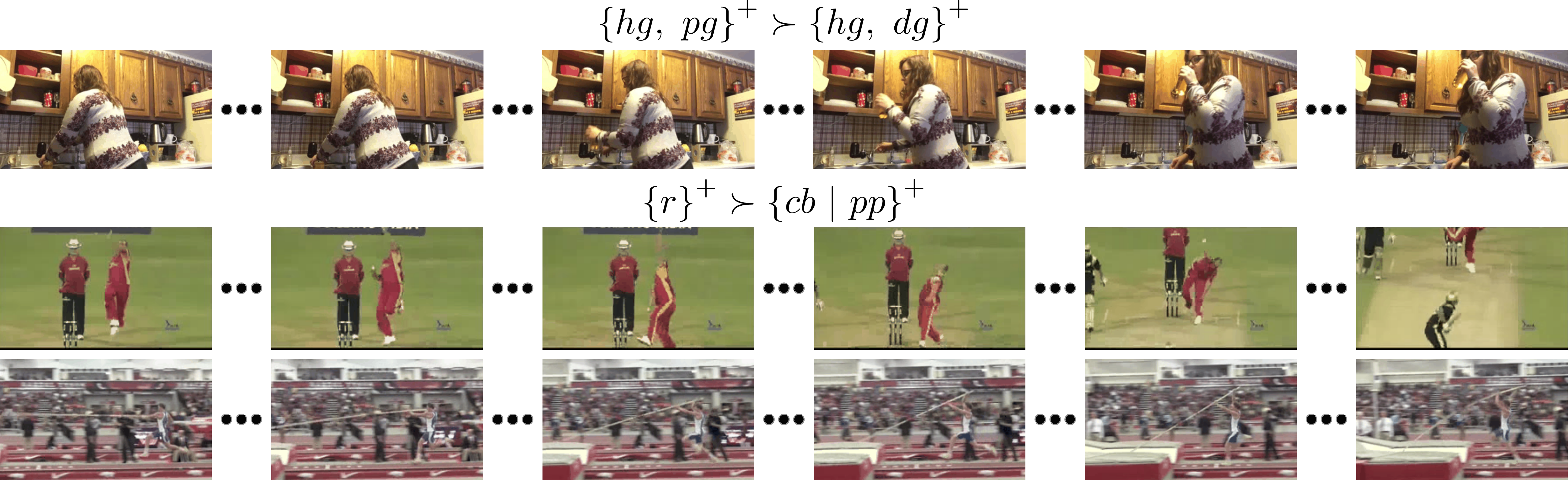}
	\end{center}
\vspace{-8px}
\caption{Examples of regular expressions and video clips matched with high probability by the proposed probabilistic inference model. The primitive actions used to form these regular expressions are holding a glass (hg), pouring water into the glass (pg), drinks from the glass (dg), running (r), cricket bowling (cb), and pole vault planting (pp).}
\label{fig:ap_vis}
\vspace{-10px}
\end{figure*}


\noindent\textbf{Trimmed Compositional Activity Classification.} We now evaluate the ability of the proposed algorithms to recognize specific activities in trimmed video clips which depict only the entire activities from the beginning to the end. Different from the previous experiment, but like the other experiments, the input regular expressions are formed by symbols that are only subsets of primitives. For instance, the symbol $\aset{\aitem{2}, \aitem{3}}$ means that the primitive actions $\aitem{2}, \aitem{3} \in \A$ happen exclusively in a frame. In addition, we mined test regular expressions with different combinations of parameters ranging jointly from 1 to 6. \tabref{table:ap_trim_apcls} presents the results.

We would like to emphasize the difficulty of the problem where the chance performance is only about 2\% MAP in both datasets. The deterministic model works only slightly better than chance, which is also a consequence of the imperfect quality of the primitive classifiers due to the difficulty of action recognition as discussed before. On the other hand, the probabilistic model provides gains above 20\% in AUC and 10\% in MAP compared to the deterministic approach in both datasets. This shows the capability of the probabilistic formulation to surpass the primitive classifiers' imprecision even when the activity of interest is very specific, producing a very complex regular expression. 
\begin{table}[h!]
\caption{Results for activity classification in trimmed videos.}
\centering
\resizebox{0.47\textwidth}{!}{\begin{tabular}{l|cc|cc|}
              & \multicolumn{2}{c|}{Expressions mined from MultiTHUMOS}                      & \multicolumn{2}{c|}{Expressions mined from Charades}                   \\
Method        & \multicolumn{1}{c}{AUC} & \multicolumn{1}{c|}{MAP} & \multicolumn{1}{c}{AUC} & \multicolumn{1}{c|}{MAP} \\ \hline
Chance        & 50.00 ($\pm$0.0) & 2.00 ($\pm 0.00$) &  50.00 ($\pm$0.0) &  2.00 ($\pm 0.00$) \\
Deterministic & 52.46 ($\pm$0.77) & 3.66 ($\pm 0.48$) &  51.85 ($\pm$0.83) &  4.40 ($\pm 1.15$) \\
Probabilistic & 73.84 ($\pm$2.63) & 13.76 ($\pm 1.93$) &  74.73 ($\pm$2.35) &  15.19 ($\pm 1.09$) \\
\hline
\end{tabular}}
\label{table:ap_trim_apcls}
\end{table}

\noindent\textbf{Untrimmed Compositional Activity Classification.} In this task, we evaluate the capability of the proposed models for recognizing specific activities in untrimmed videos which may depict the entire activity of interest at any part of the video. Here, videos can contain more than one activity, and typically large time periods are not related to any activity of interest. In this context, we modify the mined regular expressions to allow matches starting at any position in the input video. It is easily accomplished by doing the following transformation: $re \rightarrow .^\kstar re .^\kstar$ where $(.)$ is the ``wildcard'' in standard regular expression engines and in our formulation consists of every subset of primitive actions. In addition, we do not trim the video clips, instead we compute matches between the mined regular expressions and the whole video aiming to find at least one occurrence of the pattern in the entire video. We present the results on \tabref{table:ap_untrim_apcls} where we compute matches between regular expressions and the videos that have at least one positive video clip for the set of mined regular expressions.

In the same fashion as the previous experiments, the probabilistic model performs significantly better than the deterministic model. More specifically, the performance of the probabilistic model is at least 10\% better than the deterministic model in this experiment on both metrics and datasets. Therefore, the proposed probabilistic model is able to analyze entire videos and generate their global classification as accurately as it does with trimmed video clips. 

\begin{table}[h!]
\caption{Results for activity classification in untrimmed videos.}
\centering
\resizebox{0.47\textwidth}{!}{\begin{tabular}{l|cc|cc|}
              & \multicolumn{2}{c|}{Expressions mined from MultiTHUMOS}                      & \multicolumn{2}{c|}{Expressions mined from Charades}                   \\
Method        & \multicolumn{1}{c}{AUC} & \multicolumn{1}{c|}{MAP} & \multicolumn{1}{c}{AUC} & \multicolumn{1}{c|}{MAP} \\ \hline
Chance        & 50.00($\pm$0.0)  & 4.21($\pm$0.20) & 50.00($\pm$0.0) &  2.58($\pm$0.01)  \\
Deterministic & 65.69($\pm$1.34)  & 12.59($\pm$1.32) & 55.76($\pm$1.21) &  6.77($\pm$1.20)  \\
Probabilistic & 75.96($\pm$1.49)  & 26.03($\pm$1.45) & 75.43($\pm$1.35) &  17.90($\pm$1.25)  \\ \hline
\end{tabular}}
\label{table:ap_untrim_apcls}
\end{table}

\noindent\textbf{Qualitative Evaluation.} In \figref{fig:ap_vis}, we display examples of regular expressions and matched video clips using the proposed probabilistic model. In the first row, we see examples of concurrent and sequential actions where the woman depicted is holding a glass (hg) and pouring water into the glass (pg) simultaneously, and then she drinks from the glass (dg) while holding the glass. In the last two rows, we see an example of alternated actions where the desired action pattern starts with running (r) and finishes with someone either bowling (cb) or pole vault planting (pp).

\section{Conclusion}
In this paper, we addressed the problem of recognizing complex compositional activities in videos. To this end, we describe activities unambiguously as regular expressions of simple primitive actions and derive deterministic and probabilistic frameworks to recognize instances of these regular expressions in videos. Through a variety of experiments using synthetic data, we showed that our probabilistic framework excels in this task even when using noisy primitive classifiers. In the action recognition context, the proposed model was able to extend state-of-the-art action classifiers to vastly more complex activities without additional data annotation effort or large performance degradation.


\small{
\smallskip \noindent \textbf{Acknowledgements:} This research was supported by the Australian Research Council Centre of Excellence for Robotic Vision (CE140100016).}

{\small
\bibliographystyle{natbib_ieee_fullname}
\bibliography{short,ap_bib}

\begin{thebibliography}{10}\itemsep=-1pt

\bibitem[Carreira and Zisserman(2017)]{Carreira:CVPR2017}
Joao Carreira and Andrew Zisserman.
\newblock Quo vadis, action recognition? a new model and the kinetics dataset.
\newblock In {\em CVPR}, 2017).

\bibitem[El-Nouby et~al.(2016)]{Nouby:PhDThesis16}
Alaa El-Nouby, Ali Fatouh, Ahmed Ezz, Adham Gad, Ahmed Anwer, and Ehab
  Albadawy.
\newblock {\em Smart Airport Surveillance System (Action Recognition,
  Unattended Object Detection, Tracking)}.
\newblock PhD thesis, 07 2016).

\bibitem[Gao et~al.(2017)]{Gao:ICCV2017}
Jiyang Gao, Chen Sun, Zhenheng Yang, and Ram Nevatia.
\newblock Tall: Temporal activity localization via language query.
\newblock {\em ICCV}, 2017).

\bibitem[Gavrilyuk et~al.(2018)]{Gavrilyuk:CVPR2018}
Kirill Gavrilyuk, Amir Ghodrati, Zhenyang Li, and Cees~GM Snoek.
\newblock Actor and action video segmentation from a sentence.
\newblock In {\em CVPR}, 2018).

\bibitem[Habibian et~al.(2017)]{Habibian:PAMI2017}
Amirhossein Habibian, Thomas Mensink, and Cees~GM Snoek.
\newblock Video2vec embeddings recognize events when examples are scarce.
\newblock {\em PAMI}, 39(10):2089--2103, 2017).

\bibitem[Hendricks et~al.(2017)]{Hendricks:ICCV2017}
Lisa~Anne Hendricks, Oliver Wang, Eli Shechtman, Josef Sivic, Trevor Darrell,
  and Bryan Russell.
\newblock Localizing moments in video with natural language.
\newblock In {\em ICCV}, 2017).

\bibitem[Herath et~al.(2017)]{Herath:2017}
Samitha Herath, Mehrtash Harandi, and Fatih Porikli.
\newblock Going deeper into action recognition: A survey.
\newblock {\em Image and vision computing}, 60:4--21, 2017).

\bibitem[Hopcroft(1971)]{Hopcroft:1971}
John Hopcroft.
\newblock An n log n algorithm for minimizing states in a finite automaton.
\newblock {\em Theory of machines and computations}, pages 189--196, 1971).

\bibitem[Hussein et~al.(2019)]{Hussein:CVPR19}
Noureldien Hussein, Efstratios Gavves, and Arnold~WM Smeulders.
\newblock Timeception for complex action recognition.
\newblock In {\em CVPR}, 2019).

\bibitem[Hussein et~al.(2020)]{Hussein:Arxiv20}
Noureldien Hussein, Efstratios Gavves, and Arnold~WM Smeulders.
\newblock Pic: Permutation invariant convolution for recognizing long-range
  activities.
\newblock {\em arXiv preprint arXiv:2003.08275}, 2020).

\bibitem[{\.I}kizler and Forsyth(2008)]{Ikizler:IJCV2008}
Nazl{\i} {\.I}kizler and David~A Forsyth.
\newblock Searching for complex human activities with no visual examples.
\newblock {\em IJCV}, 80(3):337--357, 2008).

\bibitem[Ilie and Yu(2002)]{Ilie:2002}
Lucian Ilie and Sheng Yu.
\newblock Constructing nfas by optimal use of positions in regular expressions.
\newblock In {\em Annual Symposium on Combinatorial Pattern Matching}, 2002).

\bibitem[Jain et~al.(2015)]{Jain:ICCV2015}
Mihir Jain, Jan~C van Gemert, Thomas Mensink, and Cees~GM Snoek.
\newblock Objects2action: Classifying and localizing actions without any video
  example.
\newblock In {\em ICCV}, 2015).

\bibitem[Ji et~al.(2019)]{Ji:Arxiv19}
Jingwei Ji, Ranjay Krishna, Li Fei-Fei, and Juan~Carlos Niebles.
\newblock Action genome: Actions as composition of spatio-temporal scene
  graphs.
\newblock {\em arXiv preprint arXiv:1912.06992}, 2019).

\bibitem[Kang and Wildes(2016)]{Kang:2016}
Soo~Min Kang and Richard~P Wildes.
\newblock Review of action recognition and detection methods.
\newblock {\em arXiv preprint arXiv:1610.06906}, 2016).

\bibitem[Lampert et~al.(2014)]{Lampert:PAMI2014}
Christoph~H Lampert, Hannes Nickisch, and Stefan Harmeling.
\newblock Attribute-based classification for zero-shot visual object
  categorization.
\newblock {\em PAMI}, 36(3):453--465, 2014).

\bibitem[Lau et~al.(2010)]{Lau:2010}
Sian~Lun Lau, Immanuel K{\"o}nig, Klaus David, Baback Parandian, Christine
  Carius-D{\"u}ssel, and Martin Schultz.
\newblock Supporting patient monitoring using activity recognition with a
  smartphone.
\newblock In {\em International symposium on wireless communication systems}.
  IEEE, 2010).

\bibitem[Lawson(2003)]{Lawson:2003}
Mark~V Lawson.
\newblock {\em Finite automata}.
\newblock Chapman and Hall/CRC, 2003).

\bibitem[Liciotti et~al.(2014)]{Liciotti:2014}
Daniele Liciotti, Marco Contigiani, Emanuele Frontoni, Adriano Mancini, Primo
  Zingaretti, and Valerio Placidi.
\newblock Shopper analytics: A customer activity recognition system using a
  distributed rgb-d camera network.
\newblock In {\em International workshop on video analytics for audience
  measurement in retail and digital signage}, 2014).

\bibitem[Liu et~al.(2018)]{Liu:ECCV18}
Bingbin Liu, Serena Yeung, Edward Chou, De-An Huang, Li Fei-Fei, and
  Juan~Carlos Niebles.
\newblock Temporal modular networks for retrieving complex compositional
  activities in videos.
\newblock In {\em ECCV}, 2018).

\bibitem[Liu et~al.(2011)]{Liu:CVPR2011}
Jingen Liu, Benjamin Kuipers, and Silvio Savarese.
\newblock Recognizing human actions by attributes.
\newblock In {\em CVPR}, 2011).

\bibitem[McCulloch and Pitts(1943)]{McCulloch:1943}
Warren~S McCulloch and Walter Pitts.
\newblock A logical calculus of the ideas immanent in nervous activity.
\newblock {\em The bulletin of mathematical biophysics}, 5(4):115--133, 1943).

\bibitem[Mettes and Snoek(2017)]{Mettes:ICCV2017}
Pascal Mettes and Cees~GM Snoek.
\newblock Spatial-aware object embeddings for zero-shot localization and
  classification of actions.
\newblock In {\em ICCV}, 2017).

\bibitem[Mitkov(2003)]{Mitkov:2003}
Ruslan Mitkov.
\newblock {\em The Oxford Handbook of Computational Linguistics (Oxford
  Handbooks in Linguistics S.)}.
\newblock Oxford University Press, Inc., New York, NY, USA, 2003).

\bibitem[Piergiovanni and Ryoo(2018)]{Pier:CVPR18}
AJ Piergiovanni and Michael~S Ryoo.
\newblock Learning latent super-events to detect multiple activities in videos.
\newblock In {\em CVPR}, 2018).

\bibitem[Rabin(1963)]{Rabin:1963}
Michael~O Rabin.
\newblock Probabilistic automata.
\newblock {\em Information and control}, 6(3):230--245, 1963).

\bibitem[Rabin and Scott(1959)]{Rabin:1959}
Michael~O Rabin and Dana Scott.
\newblock Finite automata and their decision problems.
\newblock {\em IBM journal of research and development}, 3(2):114--125, 1959).

\bibitem[Richard and Gall(2016)]{Richard:CVPR2016}
Alexander Richard and Juergen Gall.
\newblock Temporal action detection using a statistical language model.
\newblock In {\em CVPR}, 2016).

\bibitem[Sedgewick and Wayne(2011)]{Sedgewick:2011}
Robert Sedgewick and Kevin Wayne.
\newblock {\em Algorithms}.
\newblock Addison-Wesley Professional, 2011).

\bibitem[Sigurdsson et~al.(2016)]{Sigurdsson:ECCV2016}
Gunnar~A. Sigurdsson, G{\"u}l Varol, Xiaolong Wang, Ali Farhadi, Ivan Laptev,
  and Abhinav Gupta.
\newblock Hollywood in homes: Crowdsourcing data collection for activity
  understanding.
\newblock In {\em ECCV}, 2016).

\bibitem[Srivastava et~al.(2015)]{Srivastava:ICML2015}
Nitish Srivastava, Elman Mansimov, and Ruslan Salakhudinov.
\newblock Unsupervised learning of video representations using lstms.
\newblock In {\em ICML}, 2015).

\bibitem[Tran et~al.(2015)]{Tran:ICCV15}
Du Tran, Lubomir Bourdev, Rob Fergus, Lorenzo Torresani, and Manohar Paluri.
\newblock Learning spatiotemporal features with 3d convolutional networks.
\newblock In {\em ICCV}, pages 4489--4497, 2015).

\bibitem[Vo and Bobick(2014)]{Vo:CVPR2014}
Nam~N Vo and Aaron~F Bobick.
\newblock From stochastic grammar to bayes network: Probabilistic parsing of
  complex activity.
\newblock In {\em CVPR}, pages 2641--2648, 2014).

\bibitem[Wang et~al.(2018)]{Wang:CVPR18}
Xiaolong Wang, Ross Girshick, Abhinav Gupta, and Kaiming He.
\newblock Non-local neural networks.
\newblock In {\em CVPR}, 2018).

\bibitem[Yeung et~al.(2017)]{Yeung:IJCV2015}
Serena Yeung, Olga Russakovsky, Ning Jin, Mykhaylo Andriluka, Greg Mori, and Li
  Fei-Fei.
\newblock Every moment counts: Dense detailed labeling of actions in complex
  videos.
\newblock {\em IJCV}, 2017).

\end{thebibliography}
}
\end{document}